\documentclass[a4paper,fleqn]{cas-dc}

\usepackage[numbers]{natbib}
\usepackage{xurl}
\usepackage{siunitx}
\usepackage{currency}
\usepackage{csquotes}
\usepackage{threeparttable}
\usepackage{bm}
\usepackage{isomath}
\usepackage[svgnames]{xcolor}
\usepackage{subcaption}
\usepackage{afterpage}
\usepackage{capt-of}
\usepackage[ruled, lined, linesnumbered]{algorithm2e} %
\usepackage{caption}
\usepackage{lipsum}
\usepackage{cuted}

\DeclareMathOperator*{\expect}{\mathlarger{\mathbb{E}}}
\renewcommand{\sectionautorefname}{Section}

\ExplSyntaxOn
\cs_gset:Npn \__first_footerline:
{
  \group_begin:
  \small\sffamily
  \textcolor{black}{
    Published~in~\textit{Software~Impacts},~Vol.~27,~2026,~100829.~
    DOI:~\href{https://doi.org/10.1016/j.simpa.2026.100829}{10.1016/j.simpa.2026.100829}
  }
  \group_end:
}
\ExplSyntaxOff

\begin{document}
\let\WriteBookmarks\relax
\def\floatpagepagefraction{1}
\def\textpagefraction{.001}
\shorttitle{Gym-TORAX: Open-source software for integrating RL with plasma control simulators}
\shortauthors{A. Mouchamps et A. Malherbe et A. Bolland et D. Ernst}

\title [mode = title]{Gym-TORAX: Open-source software for integrating reinforcement learning with plasma control simulators in tokamak research}

\author[1]{Antoine Mouchamps}
\credit{Conceptualization, Methodology, Software, Validation, Formal analysis, Writing - Original Draft, Visualization}

\author[1]{Arthur Malherbe}
\credit{Conceptualization, Methodology, Software, Validation, Formal analysis, Writing - Original Draft, Visualization}

\author[1]{Adrien Bolland}
\credit{Methodology, Validation, Formal analysis, Writing - review \& editing}

\author[1]{Damien Ernst}
\credit{Conceptualization, Writing - review \& editing, Project administration, Funding acquisition}

\affiliation[1]{organization={Montefiore Institute, University of Liège},
                city={Liège},
                country={Belgium}}

\begin{abstract}
    This paper presents Gym-TORAX, a Python package enabling the implementation of Reinforcement Learning (RL) environments for simulating plasma dynamics and control in tokamaks. Users define succinctly a set of control actions and observations, and a control objective from which Gym-TORAX creates a Gymnasium environment that wraps TORAX for simulating the plasma dynamics. The objective is formulated through rewards depending on the simulated state of the plasma and control action to optimize specific characteristics of the plasma, such as performance and stability. The resulting environment instance is then compatible with a wide range of RL algorithms and libraries and will facilitate RL research in plasma control. In its current version, one environment is readily available, based on a ramp-up scenario of the International Thermonuclear Experimental Reactor (ITER).
\end{abstract}


\begin{codemetadata}
    Current code version & 1.0.0 \\
    Link to repository & \url{https://github.com/antoine-mouchamps/gymtorax} \\
    Legal Code License & MIT license (MIT) \\
    Code versioning system used & git \\
    Software code languages, tools, and services used & Python, TORAX, Gymnasium\\
    Compilation requirements, operating environments \& dependencies & Python 3.10, TORAX 1.0.3, Gymnasium 1.2\\
    Link to developer documentation/manual & \url{https://gymtorax.readthedocs.io/latest/} \\
    Support email for questions & \href{mailto:antoine.mouchamps@gmail.com}{antoine.mouchamps@gmail.com}\\
\end{codemetadata}

\maketitle

\section{Introduction}
    One of the most prominent areas of research in fusion energy is the optimization of the stability and performance of fusion reactors. A notable share of the research effort is dedicated to reactors of the tokamak configuration, a torus-shaped device where fusion conditions are achieved through magnetic confinement \cite{Artsimovich_1972}. However, the control and design of these devices has proven to be challenging, notably because of the high dimensionality of the problem and the many nonlinearities inherent to plasma control \cite{1615272}.

    In recent years, Reinforcement Learning (RL) has\linebreak emerged as a promising approach to tackle such complex control problems. In addition to successes in domains such as robotics \cite{doi1011770278364913495721,louette2024reinforcementlearningimprovedelta}, electricity markets \cite{Boukas2021,aittahar2024optimalcontrolrenewableenergy}, and power system analysis \cite{HENRY2021100092_1,HENRY2021100092_2}, RL has also been recently applied to plasma control-related problems. In particular, \cite{Degrave2022} focused on maintaining specific plasma shapes using a set of magnetic coils, \cite{pmlr-v211-char23a} focused on controlling $\beta_N$, a plasma stability and performance measure, while \cite{seo_avoiding_2024} trained agents to avoid a certain type of plasma instability. We believe that previous successful applications have been made possible partly thanks to accessible simulation tools and software frameworks that abstract away from the user the underlying physics. This allows RL researchers to focus on optimizing the control strategy.

    To support the development of RL applications in plasma control, we present a new Python package, Gym-TORAX, a Gymnasium \cite{gymnasium} wrapper around the TORAX simulator \cite{torax2024arxiv}. Gym-TORAX allows the creation and utilization of plasma control environments that are ready-to-use by RL algorithms. These environments can be used to represent various operational scenarios such as steady-state control or ramp-up/down.
    
    The remainder of this manuscript is structured as follows. First, the Gym-TORAX package is described in \autoref{sec:gymtorax}, along with requirements for the implementation of new environments. Then, the motivation and impacts of our package on plasma control research are discussed in \autoref{sec:research_opportunities}. Finally, \autoref{sec:conclusion} concludes this manuscript with some words about future improvements and functionalities.

\section{The Gym-TORAX package}\label{sec:gymtorax}

Our package relies on the TORAX simulator for plasma dynamics. This section therefore begins with an overview of its core features and assumptions. We then describe the control problems that can be simulated using TORAX. The end of this section explains the procedure to create new Gym-TORAX environments.

\subsection{TORAX description}\label{sec:torax_desc}
    TORAX is an open-source simulator written in Python and using JAX \cite{jax2018github} for fast auto-differentiation and runtime. 

    TORAX simulates the evolution of the plasma state. This plasma state is essentially composed of the ion and electron temperatures $T_{i,e}$, the ion, electron, and impurities densities $n_{i,e,imp}$, and the poloidal magnetic flux $\psi$. Secondary metrics such as the safety factor $q$, beta coefficient $\beta$, and fusion gain $Q_{gain}$, also considered part of the state, are also computed to evaluate plasma performance. By making use of axi-symmetries, the plasma is reduced to evolving along a single spatial dimension, the normalized radius of the plasma cross-section. All space-dependent variables are discretized along this dimension.

    Every TORAX simulation starts with a configuration file, which sets, among other things, the initial conditions of the plasma state and its geometry, physical equations to solve, as well as numerical and solver-related quantities. The configuration file also sets the finite time horizon of the simulation and the numerical discretization scheme (\textit{chi} for adaptive time steps or \textit{fixed} for fixed ones).
    
    In addition to the initial values of the state variables described previously, plasma-related time series are given at each simulation time step. Those time series are related to either the total current $I_p$ or the loop voltage $V_{loop}$, the ion, impurities, and effective charge numbers $Z_{i,imp,eff}$, and various energy or particle sources. From a control perspective, the variables $V_{loop}$, $I_p$, and the energy and particle sources can be regarded as control variables. By default, TORAX therefore operates as an open-loop simulator, where these inputs are predefined for the entire simulation.

\begin{figure}[!h]
        \centering
        \includegraphics[width=0.85\linewidth]{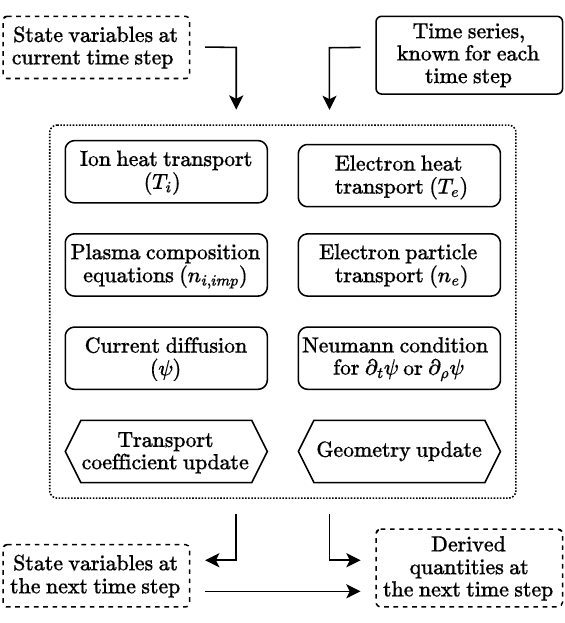}
        \caption{Update loop in a single TORAX simulation step. Given the plasma state at the current simulation time step and the time series, TORAX solves the system of equations to compute the plasma state at the next time step. Derived quantities, such as performance metrics, are then calculated from this updated state. In the figure, dashed boxes represent variables computed iteratively, while plain boxes represent time series known in advance for the whole simulation.}
        \label{fig:torax_step}
    \end{figure}

    In order to compute the plasma-state evolution over time, TORAX solves a system of  Partial Differential Equations (PDEs). Among these equations, important ones are so-called transport equations. Transport equations are PDEs that describe the spatial evolution of a given quantity in a moving medium over time. In TORAX, the transport equations related to plasma dynamics are the ion and electron heat transport and electron particle transport equations. In addition to these, other important equations are the plasma composition equations and the current diffusion. From these equations and given initial conditions and time series, the plasma state can be fully determined from one simulation time step to the next. In addition to computing the plasma state, geometric and transport equations related quantities are also updated at every simulation time step. \autoref{fig:torax_step} illustrates this update step, grouping variables into \textit{state variables}, \textit{time series}, and \textit{derived quantities} for clarity.

\subsection{Modeling approach}\label{sec:modelling}
    The Gym-TORAX package wraps around the TORAX simulator to implement a closed-loop control environment.

    The control problem is modeled as a finite-time deterministic Markov Decision Process (MDP) $(\mathcal{S}, \mathcal{A}, f, r, s_0, \gamma, T)$ with state space $\mathcal{S}$, action space $\mathcal{A}$, deterministic transition function ${f:\mathcal{S}\times\mathcal{A}\rightarrow\mathcal{S}}$, deterministic reward function ${r:\mathcal{S}\times\mathcal{A}\rightarrow\mathbb{R}}$, initial state $s_0\in\mathcal{S}$, discount factor $\gamma\in[0,\,1]$, and time horizon $T\in\mathbb{N}^+_0$.

    To implement an MDP based on the TORAX simulator, we introduce a two-level discretization. The first level is the temporal discretization corresponding to the reinforcement learning interaction cycle: at time $t$, the agent receives a complete observation of the plasma state ${s_t\in\mathcal{S}}$. From $s_t$, the agent determines the action ${a_t\in\mathcal{A}}$ to apply to the environment. The transition function $f$ applies the action and returns the next state ${s_{t+1}\in\mathcal{S}}$. An application-dependent reward ${r_t\in\mathbb{R}}$ is afterward computed.

    The second level of discretization concerns the evaluation of the transition function using TORAX. Each transition corresponds to solving the previously discussed PDEs for $K$ time-steps with TORAX. The final state of each TORAX simulation is returned as $s_{t+1}$ to the environment. In our package, two numerical discretization options are available: \textit{auto} and \textit{fixed}. Using the \textit{auto} option, the total number of simulation time steps $K_t$ within an environment transition from $t$ to $t+1$ will be dynamic and automatically defined by TORAX. On the contrary, the \textit{fixed} option enforces a constant number of time steps $K$ for each simulation run.

    An agent selects an action based on its policy $\pi$. A policy is a distribution of actions, usually conditioned on the Markov state. The objective of RL is to maximize its expected return, given by 
    \begin{equation}\label{eq:exp_return}
        J(\pi)=\expect_\pi\left[\sum_{t=0}^T \gamma^t r(s_t,a_t)\right] \quad .
    \end{equation}
    The reward function is task-specific (e.g., stability, power generation, etc.) and is designed by the user such that the expected return is maximized when the task is achieved.

    All variables given as time series in TORAX can be used as actions by the agent and shall then not be specified beforehand except for initial conditions. Those not controlled by the agent must still be specified as time series over the entire time horizon of the environment.

\subsection{Design a Gym-TORAX environment}\label{sec:design_env}
    New environments are created by extending the \verb|BaseEnv| class. In order to complete the definition of the MDP, the four following abstract methods of the \verb|BaseEnv| class have to be implemented:
    
    \begin{itemize}
        \item \verb|_get_torax_config()| defines the simulation discretization (see \autoref{sec:modelling}) and provides the TORAX configuration file (e.g., initial conditions, physical models, etc.) from which are extracted the initial state $s_0$ and the time horizon $T$.
        \item \verb|_define_action_space()| specifies the subset of TORAX actions that are controlled by the agent and optional ramp rate limits on these actions from one time step to the next;
        \item \verb|_define_observation_space()| selects which TORAX variables and derived quantities are included in the agent observation, potentially making the MDP partially observable;
        \item \verb|_compute_reward()| defines the reward function $r$ of the environment.
    \end{itemize}
    
    To simplify environment design, the \verb|_define_action_space()| and \verb|_define_observation_space()| methods rely on configurable abstract classes: the \verb|Action| class for action definitions (to specify which variable is controllable and how it relates to the TORAX configuration file) and the \verb|Observation| class for observation content. Multiple helper methods are also available in a dedicated \verb|reward| file for reward definitions. 

    In case the TORAX simulation returns an error or the state becomes unfeasible, the simulation terminates. Consequently, the episode is terminated using the dedicated Gymnasium flag, and a large negative reward ${r_t = -1000}$ is returned to the agent. If the action $a_t$ falls outside the action space $\mathcal{A}$ or if the (optional) ramp rate constraints are violated, the action is clipped before being applied to the environment, and a flag is raised in the \verb|info| return dictionary of the Gymnasium \verb|step()| function.

\section{Motivation and software impact}\label{sec:research_opportunities}

Although several plasma simulators are available to study fusion reactors, many of them are not openly accessible and require restrictive licenses, such as RAPTOR \cite{item_863d0db742814f498624d08076f58e66} or JOREK \cite{Hoelzl_2021}. Other simulators, such as EFIT \cite{Lao_1985}, are widely used for equilibrium reconstruction but do not provide dynamic plasma analyses, which makes them unsuitable for closed-loop control studies. More broadly, all of these simulators have been primarily designed for plasma physicists, which makes them hard to use by RL researchers who do not have much related expertise. Additionally, none of them provides any interface for control-oriented applications. Although TORAX \cite{torax2024arxiv} is open-source and lightweight, it suffers from the same limitations regarding control applications.

Gym-TORAX addresses these limitations by building upon TORAX to obtain a closed-loop, RL-compatible, and open-source framework. By encapsulating the plasma physics behind a classical Gymnasium API, our package lowers the entry barrier for RL researchers interested in plasma control. In this way, it fosters collaborations between both communities, allowing each one to focus on their respective expertise.

Looking forward, Gym-TORAX enables the study of new plasma control strategies and, with future developments of the software, new plasma configurations. From an RL standpoint, it allows to discover or study algorithms that are well-suited for plasma control-related problems. Even though the hypotheses inherent to the TORAX plasma simulator limit its use to preliminary investigations, its simplicity and fast execution make it a suitable starting point for advanced studies.

To support these directions, Gym-TORAX already includes a fully implemented environment based on the ITER hybrid ramp-up scenario, described in \autoref{sec:appendix_example}. Although the reward function is simple, it is ready to be used for RL training, and results can already be compared to the baselines presented in \autoref{sec:results}. We encourage users to test or improve this environment and to create new ones.

\section{Conclusion}\label{sec:conclusion}

In this paper, we introduced Gym-TORAX, an open-source framework to create Gymnasium-compatible reinforcement learning environments based on the TORAX plasma simulator. By abstracting the underlying plasma physics, Gym-TORAX enables reinforcement learning researchers to design and evaluate control strategies without requiring expert knowledge in fusion science. At the same time, it provides plasma physicists with a flexible way to implement and test control scenarios.

Future developments of Gym-TORAX will focus on expanding its flexibility and scope. A first step will be to provide tools to parameterize the plasma and tokamak geometry directly at environment creation, introducing a new dimension to the reinforcement learning problem. In addition, dedicated utilities will be added to handle specific physics-related events, such as the timing of the so-called \textit{LH} transition (the transition between \textit{L} and \textit{H} mode, which are two distinct confinement regimes; see \autoref{sec:appendix_example} for more details), which plays a critical role in plasma dynamics. Additionally, any new features extending the capabilities of the TORAX simulator can enhance the capabilities of Gym-TORAX itself.

\printcredits

\section*{Declaration of generative AI and AI-assisted technologies in the manuscript preparation process}

During the preparation of this work the authors used ChatGPT in order to correct and improve the readability, grammar, and spelling of the manuscript. After using this tool/service, the authors reviewed and edited the content as needed and take full responsibility for the content of the published article.

\bibliographystyle{elsarticle-num-names}

\bibliography{cas-refs}
\newpage
\section*{Appendix}
\appendix

\section{Package illustration}\label{sec:appendix_example}

To illustrate the capabilities of our package, in this appendix we compare the performance of three different policies in a custom environment based on an example configuration file provided in TORAX.

This appendix is organized into three parts. First, a description of the simulation environment is given in \autoref{sec:env_desc}. Then, the three policies that will be evaluated are presented in \autoref{sec:agents}. Finally, \autoref{sec:results} comments on the results.

\subsection{Environment description}\label{sec:env_desc}

The environment used for this case study is the ITER hybrid ramp-up scenario, provided as an example configuration file in TORAX. This scenario is based on the work of \cite{Citrin_2010} and consists of a power ramp-up phase of 100 seconds, followed by a nominal phase lasting 50 seconds. The first phase of 100 seconds takes place in so-called \textit{L-mode} (low-confinement regime), while the nominal phase occurs in \textit{H-mode} (high-confinement regime). These are two distinct plasma confinement regimes with different physical properties.

The environment is named \verb|IterHybridEnv|. The environment features three actions: \verb|IpAction| for the total current, \verb|NbiAction| for NBI (Neutral Beam Injection), and \verb|EcrhAction| for ECRH (Electron-Cyclotron Resonance Heating), both representing external energy sources. Its action space $\mathcal{A}$ is bounded, and a ramp-rate limit is imposed on the total current. By default, the environment is fully-observable and uses the \verb|AllObservation| class with custom bounds applied to certain variables.

The reward function is a linear combination of four elements:
\begin{equation}\label{eq:reward}
    r = \alpha_Q\cdot g_Q + \alpha_{q_{min}}\cdot g_{q_{min}} + \alpha_{q_{95}}\cdot g_{q_{95}} + \alpha_\mathrm{H98}\cdot g_\mathrm{H98}\quad.
\end{equation}
In this equation, $\alpha_i$ and $g_i$, with ${i\in\{Q,q_{min},q_{95},\mathrm{H98}\}}$, represent weights and functions, respectively. These are related to the fusion gain $Q$, the minimum $q_{min}$ and edge $q_{95}$ safety factors, and the H-mode confinement quality factor $\mathrm{H}98$, respectively.

\subsection{Policies}\label{sec:agents}

We consider three different policies: an open-loop policy, a random policy, and one Proportional Integral (PI) controller-based policy. The open-loop policy $\pi_{OL}$ uses a predetermined set of actions that directly follows the action trajectories of the initial scenario given in TORAX. This serves as the reference scenario for the two other policies. The random policy $\pi_{R}$ selects the actions uniformly at random. The PI controller-based policy $\pi_{PI}$ controls the total current action using a PI controller and uses the same predetermined trajectories as the open-loop policy for the last two actions, NBI and ECRH. The PI controller is used to follow a prescribed linear increase (from \qty{0.6}{MA/m^2} to \qty{2}{MA/m^2}) of the current density at the center of the plasma during the ramp-up phase of 100 seconds. Once the ramp-up has been performed, action values are kept constant from the last action given by the PI controller until the end of the episode (for the last 49 seconds).

\begin{figure}[!h]
    \centering
    \includegraphics[width=1\linewidth]{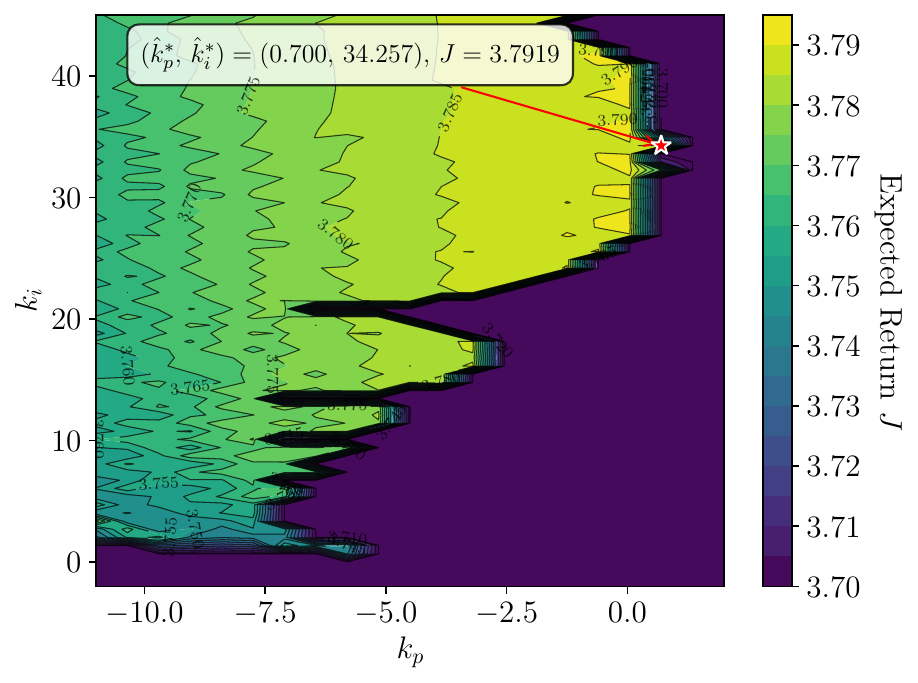}
    \caption{Heatmap of the expected return $J$ over a subset of the full parameter space. To improve the clarity of the figure, the scale used to represent $J$ values is clipped at a minimum of $3.7$.}
    \label{fig:heatmap}
\end{figure}

The proportional $k_p$ and integral $k_i$ gains of the PI controller are optimized to maximize the expected return $J(\pi)$. The optimization is carried out using a grid search. First, we define a subspace ${\mathcal{K}_p\times \mathcal{K}_i\subset \mathbb{R}^2}$ of the full parameter space. This subspace is determined so as to avoid regions of large reward penalties and regions where the expected return does not evolve anymore. The parameter space ${\mathcal{K}_p\times \mathcal{K}_i}$ is then discretized into $n\cdot m$ discrete points by using $n$ and $m$ values for $k_p$ and $k_i$, respectively. An approximation of the optimal parameters can then be computed by selecting the point, among all, whose corresponding policy maximizes the expected return. Running this algorithm using ${n=20}$ and ${m=60}$ as hyperparameters, we obtain the expected return topology depicted in \autoref{fig:heatmap}, from which we obtain the following estimations for the optimal parameters: ${\hat{k}_p^*=0.700}$ and ${\hat{k}_i^*=34.257}$.

\subsection{Results}\label{sec:results}

The expected return defined in \autoref{eq:exp_return} obtained for each policy is given in \autoref{tab:results}.

\begin{table}[width=.9\columnwidth,cols=2,pos=h]
    \begin{tabular*}{\tblwidth}{@{} LR@{} }
        \toprule
        Policy & Expected return \\
        \midrule
        $\pi_{OL}$ & $3.40$ \\
        $\pi_R$ & $-10.79$ \\
        $\pi_{PI}$ & $3.79$ \\
        \bottomrule
    \end{tabular*}
    \caption{Expected return of the three studied policies, using a discount factor ${\gamma=1}$.\label{tab:results}}
\end{table}

The open-loop policy yields an average return of $3.40$. As expected, the random policy performs worse, with an average return of $-10.79$. The best-performing policy is the PI controller-based one, with an average return of $3.79$. This result is an improvement over the reference scenario and can serve as a baseline for more sophisticated policies.

\begin{figure}[!h]
    \centering
    \includegraphics[width=0.6\linewidth]{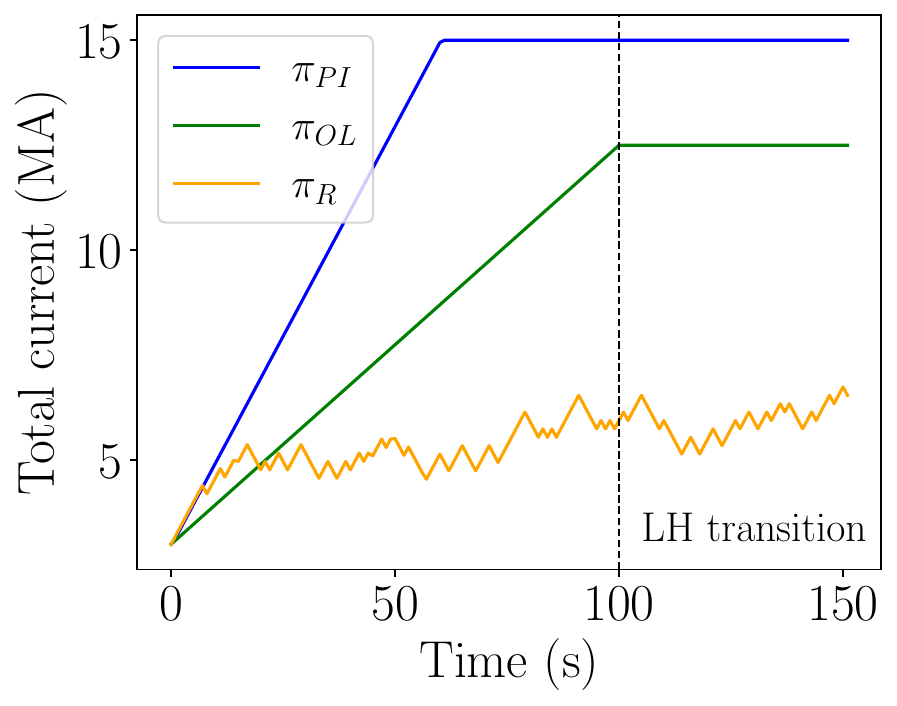}
  \caption{Comparison of one action (total current) trajectory for each policy.}\label{fig:total_current}
\end{figure}

A representation of an action (total current) trajectory for each policy is given in \autoref{fig:total_current}. This figure shows the erratic evolution of the total current of the random policy, which is somewhat mitigated by the ramp-rate constraints imposed in the environment. Regarding the PI policy, the trajectory of the current increases steadily and levels off at $\qty{15}{MA}$, the maximum value allowable in the environment. This behavior is consistent with the fact that higher values of total current can generally be associated with improved confinement and overall better performance \cite{ITER_Physics_Expert_Group_on_Confin_Transport1999-hu}.

\begin{figure}[!h]
    \centering
    \includegraphics[width=0.6\linewidth]{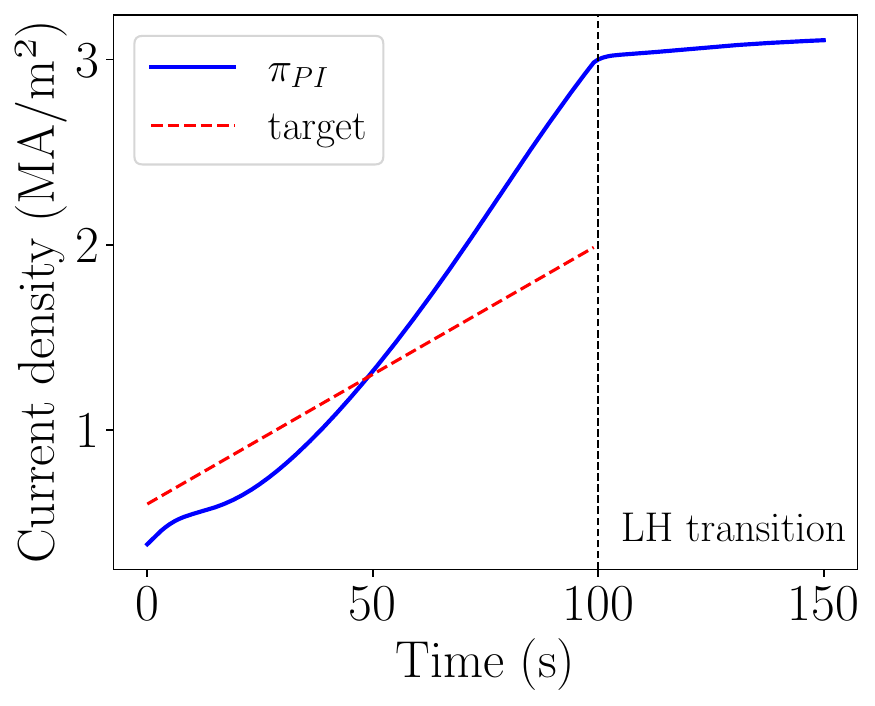}
  \caption{Evolution of the current density with respect to the target.}\label{fig:error}
\end{figure}

\autoref{fig:error} represents the target evolution of the PI controller and the action taken by the PI controller-based policy. Note that the parameters were optimized to maximize the expected return, rather than having actions close to the target, which can be observed in the figure.

\end{document}